\pdfoutput=1

\documentclass[11pt]{article}

\usepackage{emnlp2021}

\usepackage{times}
\usepackage{latexsym}

\usepackage[T1]{fontenc}

\usepackage[utf8]{inputenc}

\usepackage{microtype}

\usepackage{svg}
\usepackage{float}
\usepackage{todonotes}
\usepackage{hyperref}
\usepackage{graphicx}
\usepackage{amsmath}
\usepackage{amssymb} 
\usepackage{booktabs}
\usepackage{xcolor}
\usepackage{enumitem}
\usepackage{float}
\usepackage{xspace}

\definecolor{tablegray}{rgb}{0.2,0.2,0.2}
\newcommand{\stdintable}[1] {~~\textcolor{tablegray}{\scriptsize{$\pm$#1}}}

\definecolor{ao}{rgb}{0.0, 0.5, 0.0}
\definecolor{tb}{cmyk}{0, 0.7808, 0.4429, 0.1412}

\title{AdapterHub Playground:\\ Simple and Flexible Few-Shot Learning with Adapters}

\author{Tilman Beck$^{1}$, Bela Bohlender$^{1}$,  Christina Viehmann$^{2}$, {\bf Vincent Hane$^{1}$}, \\ {\bf Yanik Adamson$^{1}$}, {\bf Jaber Khuri$^{1}$}, {\bf Jonas Brossmann$^{1}$}, {\bf Jonas Pfeiffer$^{1}$}, {\bf Iryna Gurevych$^{1}$} 
\\ $^{1}$Ubiquitous Knowledge Processing Lab (UKP Lab)\\ Department of Computer Science \\Technical University of Darmstadt
\\ $^{2}$Institut f{\"u}r Publizistik \\Johannes Gutenberg-University Mainz}

\begin{document}
\maketitle
\begin{abstract}
The open-access dissemination of pretrained language models through online repositories has led to a democratization of state-of-the-art natural language processing (NLP) research.
This also allows people outside of NLP to use such models and adapt them to specific use-cases.
However, a certain amount of technical proficiency is still required which is an entry barrier for users who want to apply these models to a certain task but lack the necessary knowledge or resources.
In this work, we aim to overcome this gap by providing a tool which allows researchers to leverage pretrained models without writing a single line of code.
Built upon the parameter-efficient adapter modules for transfer learning, our AdapterHub Playground provides an intuitive interface, allowing the usage of adapters for prediction, training and analysis of textual data for a variety of NLP tasks.
We present the tool's architecture and demonstrate its advantages with prototypical use-cases, where we show that predictive performance can easily be increased in a few-shot learning scenario.
Finally, we evaluate its usability in a user study.
We provide the code and a live interface\footnote{\url{https://adapter-hub.github.io/playground}}.

\end{abstract}

\section{Introduction}
\label{sec:intro}
The success of transformer-based pretrained language models~\cite{devlin-etal-2019-bert, liu2019roberta} was quickly followed by their dissemination, gaining popularity through open-access Python libraries like Huggingface~\cite{wolf-etal-2020-transformers}, AdapterHub~\cite{pfeiffer-etal-2020-adapterhub} or SBERT~\cite{reimers-gurevych-2019-sentence}.
Researchers and practitioners with a background in computer science are
able to download models and fine tune them to their needs.
They can then upload their fine-tuned model and contribute to an open-access community of state-of-the-art (SotA) language models for various tasks and in different languages.

This has significantly contributed to the democratization of access to the latest NLP research as the individual implementation process has been simplified through the provision of easy-to-use and actively managed code packages. 
However, one still needs a certain level of technical proficiency to access these repositories, train models, and predict on new data.
This is a limiting factor for researchers in disciplines who could benefit from applying SotA NLP models in their field, but lack the technical ability.
Furthermore, there is growing interest for text classification models in interdisciplinary research~\cite{van2021validity, boumans2016taking}, although often the methods are not SotA in NLP.

In this work, we hope to bridge this gap by providing an application which makes the power of pretrained language models available without writing a single line of code.
Inspired by the recent progress on parameter-efficient transfer learning~\cite{NIPS2017_e7b24b11, houlsby2019parameter}, our application is based on adapters which introduce small and learnable task-specific layers into a pretrained language model.
During training, only the newly introduced weights are updated, while the pre-trained parameters are frozen. Adapters have been successfully applied in machine translation~\citet{bapna-firat-2019-simple, philip-etal-2020-monolingual}, cross-lingual transfer~\cite{pfeiffer-etal-2020-mad, pfeiffer-etal-2021-unks, ustun-etal-2020-udapter, vidoni2020orthogonal}, community QA~\cite{ruckle-etal-2020-multicqa}, task composition for transfer learning~\cite{pmlr-v97-stickland19a, pfeiffer-etal-2021-adapterfusion, lauscher-etal-2020-common, wang-etal-2021-k} and text generation~\cite{ribeiro-etal-2021-structural}.
Adapters are additionally computationally more efficient~\cite{ruckle-etal-2021-adapterdrop} and more robust to train~\cite{he-etal-2021-effectiveness, han-etal-2021-robust}.
In our work, we build our application on top of the AdapterHub~\cite{pfeiffer-etal-2020-adapterhub} library which stores task-specific adapters with a large variety of architectures and offers upload functionalities for community-developed adapter weights.
We leverage this library to allow no-code access to pre-trained adapters for a many text classification tasks using dynamic code generation. 
Finally, our application enables the analysis of multi-dimensional annotations to further investigate model performance.

Our application supports both NLP and interdisciplinary researchers who want to evaluate the transferability of existing pretrained adapters to their specific domains and use cases.
We intend this application for both zero-shot as well as few-shot scenarios where a user annotates a small number of data points and monitors model improvements.
This is especially interesting for \textit{intermediate} task training~\cite{phang2018sentence} where models trained on a compatible task are utilized and fine-tuned on the target task.

\begin{figure}
    \centering
    \includegraphics[width = 180pt]{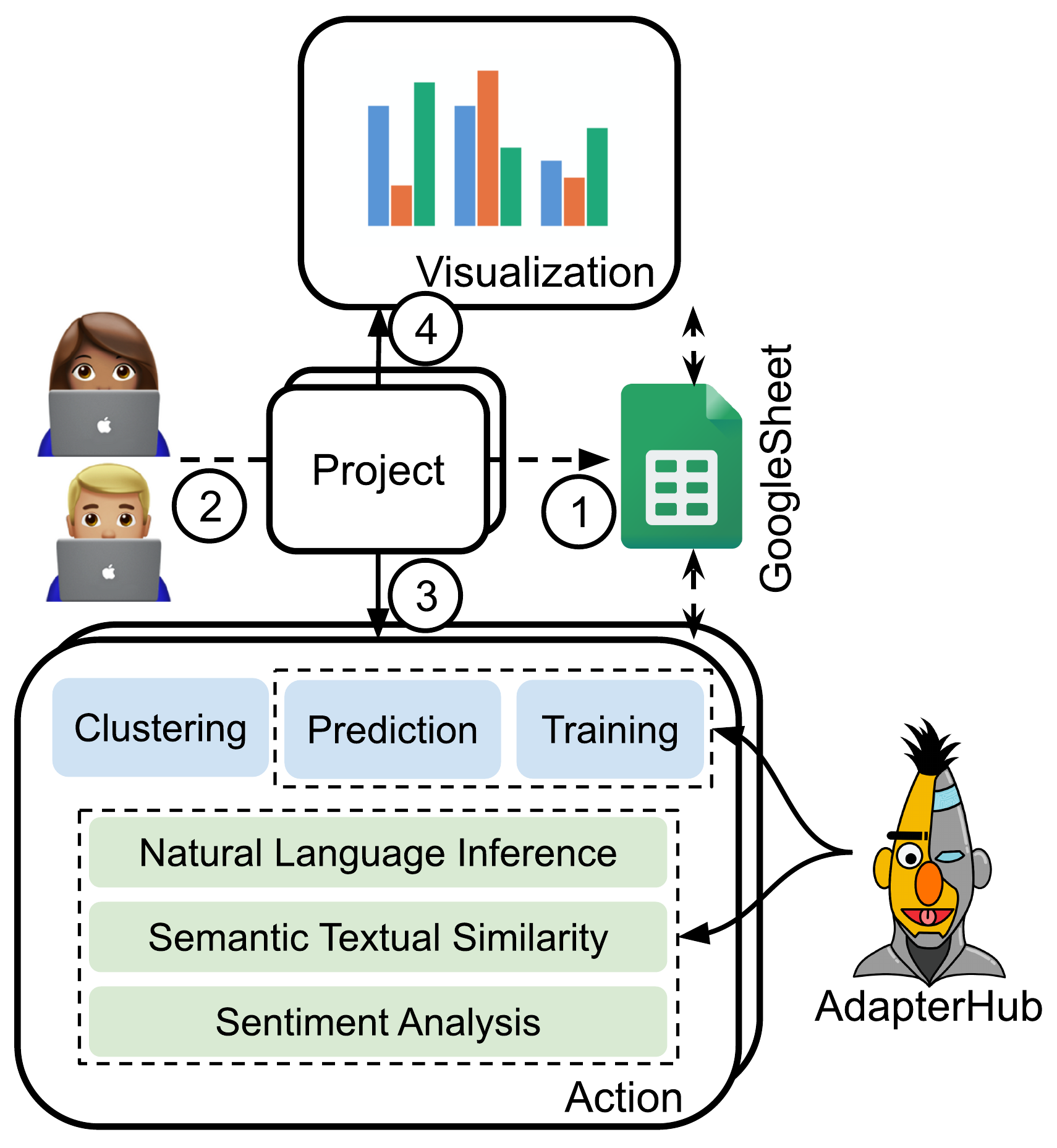}
    \caption{Diagram of the AdapterHub Playground workflow. \raisebox{.5pt}{\textcircled{\raisebox{-.9pt} {1}}} Users upload their text data to GoogleSheets  and \raisebox{.5pt}{\textcircled{\raisebox{-.9pt} {2}}} link it to a new project. \raisebox{.5pt}{\textcircled{\raisebox{-.9pt} {3}}} In each project, users can create multiple actions  by selecting a specific action type \texttt{Training}, \texttt{Prediction}, \texttt{Clustering}. For the \texttt{Training} and \texttt{Prediction} action types, the user needs to define the desired downstream task (e.g. \texttt{Sentiment Analysis}). Information about available pretrained adapters for the specified task are dynamically retrieved from AdapterHub. \raisebox{.5pt}{\textcircled{\raisebox{-.9pt} {4}}} After generating predictions, the user can visualize the results  within the project.}
    \label{fig:workflow}
    \vspace{-1em}
\end{figure}

Some efforts are being made to abstract away engineering requirements to use SotA NLP~\cite{akbik-etal-2019-flair}, but their usage still requires certain technical skills.
Existing no-code (or AutoML) applications like Akkio\footnote{\url{https://www.akkio.com/}}, Lobe\footnote{\url{https://lobe.ai/}}, or Teachable Machines\footnote{\url{https://tinyurl.com/teachablemachines}} allow users to upload data, annotate it using self-defined labels, and train a model for prediction.
Most approaches focus on vision tasks and follow commercial goals.
To the best of our knowledge, we are the first to provide a non-commercial, no-code application for text classification. 
Our application is transparent (i.e. details about usable pretrained adapters are traceable), and extendable via the community-supported AdapterHub library.
Finally, we enable execution on third party computational servers for users without access to the required GPU hardware for efficient training and prediction~\cite{ruckle2020adapterdrop}, while also providing the necessary scripts to setup a self-hosted computing instance, mitigating technical dependencies.

Our contributions are: \textbf{1)}~The AdapterHub Playground application which enables no-code inference and training by utilizing pretrained adapters; \textbf{2)}~Prototypical showcase scenarios from social sciences using our application for few-shot learning; \textbf{3)}~An elaborate user study  that analyzes the usability of our proposed application.

\section{AdapterHub Playground}
\label{sec:playground}
The AdapterHub Playground is a lightweight web application offering no-code usage of pretrained adapters from the AdapterHub library.
A user interface accompanied by dynamic code generation allows the utilization of adapters for inference and training of text classification tasks on novel data.
Below, we describe the application workflow\footnote{We provide information about user requirements in the Appendix~\ref{app:qa}.}, provide details on the specific functionalities and highlight the technical architecture.

\subsection{Workflow}

The workflow of the AdapterHub Playground is depicted in Figure~\ref{fig:workflow}. 
First, a user creates a GoogleSheet\footnote{\url{https://docs.google.com/spreadsheets/}} and uploads the input data for the desired classification task. 
If applicable, additional metadata, for example, annotations or timestamps, can be added.
Next, a new project can be created and linked to the data via the GoogleSheet sharing functionality. 
Within a project, the user can define an \textit{action}, resembling a computational unit (e.g. training an adapter).
Upon submission of a new action, the input text data is downloaded and the specified computation is performed.
The user is informed visually about the status of the execution in the application. 
After finishing the computation, the results are written directly into the GoogleSheet by the system and evaluation details are provided in the action interface.
By default the system supports accuracy and macro-F1 evaluation metrics.
To aid users in estimating model performance we additionally provide the results for random and majority prediction.

A user can create multiple projects, and within each project, multiple actions can be triggered using the same input data.\footnote{This allows direct comparison among multiple adapters.}
Finally, within a project a user can explore the predictions on the data using different visualization methods.

\begin{figure}
    \centering
    \includegraphics[scale=0.40]{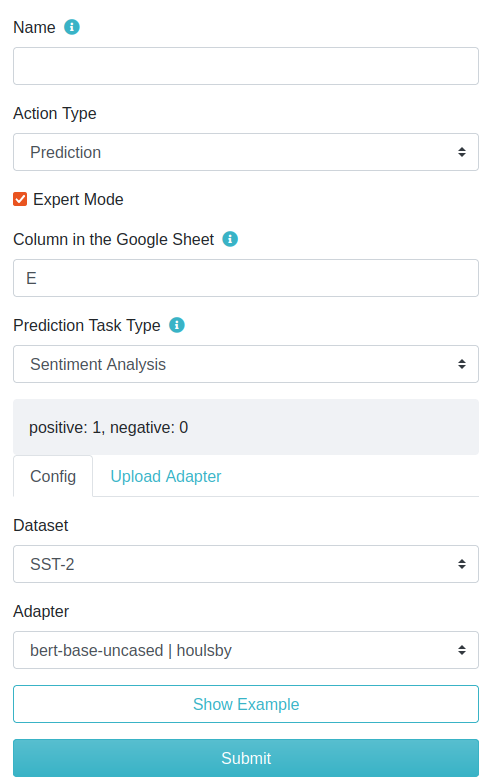}
    \caption{Screenshot of the action creation dialogue. A user has to provide a name for the action, the action type (here \texttt{Prediction}), the column in GoogleSheet where results are written to and the downstream task (here \texttt{Sentiment Analysis}). In the expert mode, the user has additional options for the pretrained adapter, i.e. the dataset which was used for pretraining and the specific architecture. The available options are dynamically retrieved from AdapterHub. Alternatively a self-provided adapter can be uploaded.}
    \label{fig:taskcreation}
    \vspace{-1em}
\end{figure}

\subsection{Actions}
\label{ssec:tasks}

Our application focuses on three main \textit{actions}, namely \texttt{Prediction}, \texttt{Training} and \texttt{Clustering}.
For each action, the respective code is dynamically generated by merging static code snippets with parameters defined by the user (e.g. the specific adapter architecture).
In the following we describe the procedure of each action.

\paragraph{Prediction.}
Pretrained task-specific adapters can be utilized for predictions on proprietary data.
The user creates a new action in the project detail page and selects as action type \texttt{Prediction}, defines the column of the GoogleSheet in which the predictions should be written, and selects the respective downstream task which is dynamically retrieved from the AdapterHub.\footnote{We currently focus on (pairwise) text classification tasks.}
Execution triggers the backend program to load the specified adapter and data, and produce task-specific labels for the data.
A screenshot of the action creation dialogue is provided in Figure~\ref{fig:taskcreation}.

\paragraph{Training.}
To allow for continual training of adapters on labeled data, the user creates a new action of type \texttt{Training}.
When executed, the backend process loads the specified adapter, downloads both data and target labels, and starts the training procedure.
Once training is completed, the user can download the fine-tuned adapters as a zipped file.
This makes fine-tuned adapter weights available for another \texttt{Prediction} action.

The choice of hyperparameters can have substantial influence on task performance but evaluating these effects is out of scope for this work.
Defaults are set based on the literature~\cite{pfeiffer-etal-2021-adapterfusion}, however, if necessary, the user can modify training hyperparameters through various dropdown fields.
This allows to compare multiple adapters trained with different hyperparameters.

\paragraph{Clustering.}
Discovering recurrent patterns in text data is a common procedure in various research disciplines.
To allow for deeper text analysis, we additionally provide the \texttt{Clustering} action which enables users to apply clustering algorithms on the data based on their textual similarity.
We provide K-Means and hierarchical clustering~\cite{scikit-learn} as algorithm choices and support Tf-Idf and SBERT embeddings~\cite{reimers-gurevych-2019-sentence} as text representations.

\begin{figure}
    \centering
    \includegraphics[width = 200pt]{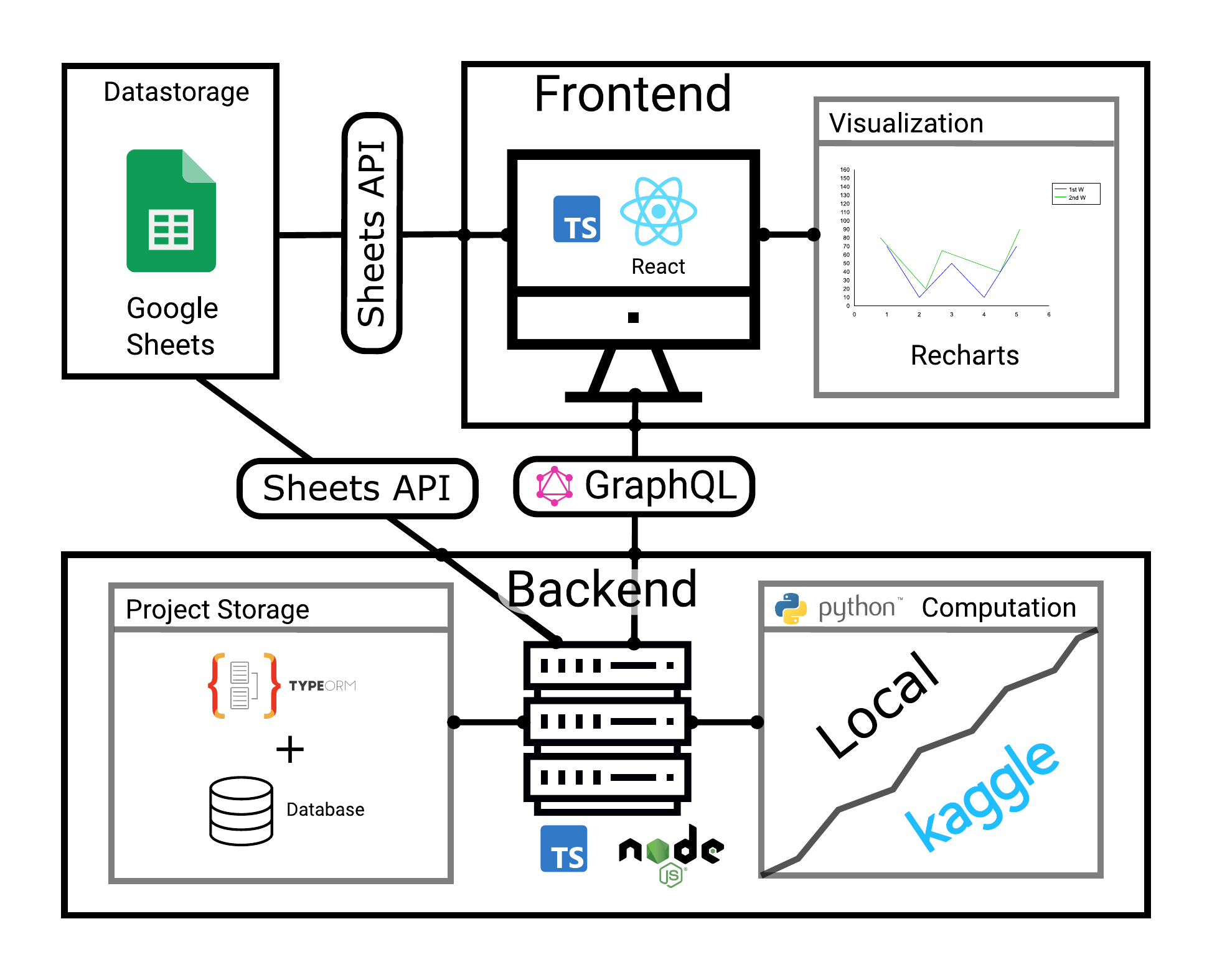}
    \caption{The AdapterHub Playground architecture}
    \label{fig:architecture}
    \vspace{-1em}
\end{figure}

\subsection{Architecture}
\label{ssec:architecture}

The architecture of the AdapterHub Playground (Figure~\ref{fig:architecture}) is designed to be easy to setup and requires a minimal set of dependencies.
The tool is based on three main components; \texttt{Frontend}, \texttt{Backend} and \texttt{Data Storage}.
A user interacts with the frontend and triggers various actions as described in \S\ref{ssec:tasks}.
The backend receives instructions via the frontend and manages the execution on the computational resource.
Our application additionally hosts a local database for user and project management.

We chose GoogleSheet as text data storage component due to its similarity to established easy-to-use spreadsheet applications.
It supports a variety of import and export mechanisms which simplify the data management process, especially for non-technical users. 
GoogleSheet also reduces storage requirements on the local computational resource, keeping the application lightweight and manageable.
Finally, the Sheets API\footnote{\url{https://tinyurl.com/SheetsAPI}} provides a programmatic interface for communication.
Although this requires users to use a Google account, we argue the advantages compensate for this restriction.

Below, we describe the technical details of the implementation for the specific components.

\vspace{1mm}
\noindent\textbf{Frontend.}
The frontend provides the visual interface for management (i.e. creation, editing, deletion) of projects and their respective actions.
After login with an authentication token\footnote{Depending on the chosen backend solution, this can be a JSON file provided by the system administrator of the backend server or the authentication token provided by Kaggle.}, a webpage lists all user projects.
By selecting a project, a corresponding details page allows actions to be managed (see \S\ref{ssec:tasks}) and visualizations to be created using project-specific data storage.

The frontend is implemented using the React\footnote{\url{https://reactjs.org/}} framework and is written in TypeScript\footnote{\url{https://www.typescriptlang.org/}}.
The frontend design is based on Bootstrap\footnote{\url{https://getbootstrap.com/}}.
Communication with the backend is realized via the GraphQL query language.
The data is retrieved using the Sheets API and can be visualized via Recharts\footnote{\url{https://recharts.org}} which offers seamless integration of the D3\footnote{\url{https://d3js.org/}} visualization library within the React framework.

\vspace{1mm}
\noindent\textbf{Backend.}
The backend organizes the storage of application-relevant objects (i.e. users, projects, tasks) and manages both dynamic code generation and execution.
User credentials, projects, and tasks are stored in a SQL database.
When an action is executed in the frontend, the backend server loads the task-specific code template and dynamically integrates parameter information provided  for the individual task.
Depending on the choice of the computational node, the generated Python script is scheduled for execution either locally or on a Kaggle compute node via the KaggleAPI\footnote{\url{https://www.kaggle.com/docs/api}}.

The backend is implemented using Node.js\footnote{\url{https://nodejs.org}} and TypeScript.
For application-relevant data, any TypeORM\footnote{\url{https://typeorm.io/}}-supported database (e.g. MySQL, PostgreSQL, etc.) can be used.
Communication with data storage is realized via Sheets API.

\section{Few-shot scenario}
\label{sec:fewshot}
Several prominent tasks in NLP such as sentiment analysis~\cite{socher-etal-2013-recursive, rosenthal-etal-2017-semeval}, stance detection~\cite{mohammad-etal-2016-dataset, schiller2021stance} or identifying semantically similar texts~\cite{cer-etal-2017-semeval, agirre-etal-2012-semeval} are of great interest in social science research~\cite{boumans2016taking, beck2021investigating, van2021computational}.
We therefore replicated two scenarios, namely sentiment analysis and semantic textual similarity.

We envision a situation where a user has collected textual data (e.g. sentence-level) for a given task and wishes to perform analysis using a text classification pipeline.
A labeled test set to evaluate the performance of the classifier, and further training data is available.

\subsection{Experiments}

\noindent\textbf{Data.}
For demonstration purposes, we recreated the above-mentioned scenario using existing datasets for both tasks. 
For sentiment analysis, we use the dataset by \citet{barbieri-etal-2020-tweeteval}.
In particular, we retrieve text for the Twitter Sentiment Analysis dataset which was originally used for the Semeval2017 Subtask A~\cite{rosenthal-etal-2017-semeval}.
At time of writing, the AdapterHub provides mostly pretrained adapters for binary sentiment classification (\textit{positive}, \textit{negative}).
Thus, we discarded all items labeled as \textit{neutral} from the dataset and are left with 24,942 Tweets for training and 6,347 Tweets for testing.

For semantic textual similarity, we use the dataset by \citet{lei-etal-2016-semi} which is a set of pairwise community questions from the AskUbuntu\footnote{\url{https://askubuntu.com/}} forum annotated for duplicates.
Specifically, we use the question titles of the human-annotated development (4k) for training and the test instances (4k) for testing.

\vspace{1mm}
\noindent\textbf{Setup.}
For binary sentiment classification, we use the AdapterHub to obtain three different adapters which were previously trained~\cite{pfeiffer-etal-2021-adapterfusion} on English datasets from the movie review domain.
The \texttt{IMDB} adapter was fine-tuned on the dataset by \citet{maas-etal-2011-learning}, the \texttt{RT} adapter was trained on the Rotten Tomatoes Movie Reviews dataset by \citet{pang-lee-2005-seeing}, and the \texttt{SST-2} adapter was trained using a binarized dataset provided by ~\citet{socher-etal-2013-recursive}.

For semantic textual similarity, we obtained the \texttt{MRPC} adapter trained on the paraphrase dataset by \citet{dolan-brockett-2005-automatically} and the \texttt{QQP} adapter trained on the Quora Duplicate Question dataset.\footnote{\url{https://tinyurl.com/quora-qp}}

The experiments were conducted using the AdapterHub Playground without writing any code.
We experiment with different training dataset sizes, repeated three times with different subsets of the training data randomly selected for each run.\footnote{See Appendix~\ref{app:training} for experimental details.}. 
We evaluated statistically significant differences ($p<0.05$) between zero-shot and few-shot results of each adapter using a paired Bootstrap test~\citep{efron1994introduction}.

\subsection{Results}

\begin{table*}[tb]
\small
    \centering
    \begin{tabular}{lcccccccc}
    \toprule
    Adapter & 0 & 5 & 10 & 20 & 50 & 100 & 1,000 & N \\ 
    \midrule
    \texttt{IMDB} & 71.99 & 65.40 \stdintable{2.08} & \textbf{72.25} \stdintable{14.78} & 67.51 \stdintable{11.25} & 71.37 \stdintable{4.93} & \underline{81.87} \stdintable{2.49} & \underline{84.10} \stdintable{5.34} & \underline{88.36} \\
 
    \texttt{RT} & 76.24 & 72.33 \stdintable{0.97} & \textbf{76.76} \stdintable{10.08} & 67.38 \stdintable{10.09} & 67.44 \stdintable{5.57} & 76.88 \stdintable{4.22} & 82.64 \stdintable{6.01} & \underline{90.50} \\

    \texttt{SST-2} & 84.61  & 84.53 \stdintable{0.15} & \textbf{86.23} \stdintable{2.91} & 84.22 \stdintable{0.53} & 82.33 \stdintable{2.41} & 83.54 \stdintable{0.61} & \underline{88.19} \stdintable{2.08} & \underline{92.04} \\

    \midrule
    
    \texttt{MRPC} & 31.18 & \textbf{31.64} \stdintable{5.37} & 29.22 \stdintable{0.08} & 28.46 \stdintable{1.30} & \underline{38.57} \stdintable{3.66} & \underline{36.66} \stdintable{8.54} & 28.31 \stdintable{3.18} & 26.23 \\

    \texttt{QQP} & 73.28 & 73.10 \stdintable{0.16} & 73.19 \stdintable{0.06} & 72.79 \stdintable{0.44} & 71.08 \stdintable{1.17} & 69.60 \stdintable{0.48} & 73.01 \stdintable{0.30} & \textbf{73.68} \\
    \bottomrule
    \end{tabular}
    \caption{Few-shot performance of various pretrained adapters from AdapterHub using increasing size of training data. Underlined scores are significantly ($p<.05$) better than their zero-shot counterpart. Bold scores resemble experiments with minimum training data required for outperforming zero-shot performance of respective adapter. All numbers are accuracy scores. N is for using all available training data.}
    \label{tab:results}
    \vspace{-1em}
\end{table*}

The results for both tasks are shown in Table~\ref{tab:results}.

\vspace{1mm}
\noindent\textbf{Sentiment Analysis.}
The overall best performance is achieved by the \texttt{SST-2} adapter, simultaneously the most robust performance in terms of the standard deviation across different runs and varying amounts of training data.
This is most likely due to the substantially larger size of the initial training data (\texttt{SST-2}: 67k, \texttt{RT}: 8k, \texttt{IMDB}: 25k) for the adapter.
Although, on average, for all adapters zero-shot performance could be outperformed using a minimum of 10 instances, the differences between individual runs vary largely and statistically significant improvements are only achieved using a larger number of training instances (e.g., at least N$\geq$100 for \texttt{SST-2}).
We find using a small number of annotated examples (N$\leq$50) leads to worse performance compared to zero-shot performance (N=0) and to less robust results across runs with randomly sampled training data.
Providing 1,000 training samples leads to significant improvements for adapters \texttt{IMDB} and \texttt{SST-2} but only providing the full dataset results in statistically significant improvements for all adapters.

\vspace{1mm}
\noindent\textbf{Semantic Textual Similarity.}
The performance gap between both adapters is large, with a difference of 42.10 in the zero-shot setting, favoring \texttt{QQP}.
The results for the \texttt{MRPC} adapter show no clear tendency to improve as the training data size grows, with performance peaking at 50 training instances.
Most surprisingly, using 1,000 or all available training samples (4k) leads to a severe performance decrease.
For the \texttt{QQP} adapter, performance variations are minimal and none of the few-shot experiment settings leads to a significant improvement over zero-shot performance.

\vspace{1mm}
\noindent\textbf{Summary.}
\citet{poth-etal-2021-pre} investigated the effects of intermediate task fine-tuning in adapter settings.
They showed that domain similarity, task type match and dataset size are good indicators for the identification of beneficial intermediate fine-tuning tasks.
Our experiments confirm this finding although we cannot observe consistent improvement with larger training data size.
Thus, more research on robust few-shot learning is necessary.

In contrast to relying on off-the-shelf tools for automated content analysis, our application enables direct evaluation of both zero-shot and few-shot performance of existing pretrained adapters.
This is especially helpful for assessment of the applicability of such models for interdisciplinary research~\cite{grimmer2013text} but can also be used to test robustness with varying hyperparameter configurations.

\section{Usability Study}
\label{sec:userstudy}
AdapterHub Playground is designed to be simple to use, requiring minimal training effort and technical knowledge.
While we followed these principles throughout the conception and implementation of the application, we also evaluated the usability with users from our target group.
Therefore, we followed the approach by \citet{hultman2018usability} and let study participants conduct a series of tasks which were designed to reflect a use-case scenario as described in \S\ref{sec:fewshot}.
Afterwards, we used a questionnaire to capture their experiences.

\vspace{1mm}
\noindent\textbf{Participants.}
We recruited study participants (N=11) from the communication science field, the majority of whom were (post)graduate-level researchers at a university (two Professors, two PostDocs, six PhDs, one B.Sc.).
Our data suggests that the participants have limited or no understanding of the technical computer science concepts but can envision themselves using the AdapterHub Playground (for details see Appendix~\ref{app:usability}).
Thus, our participants belong to one of the target groups we aim to aid with this application.

\vspace{1mm}
\noindent\textbf{Procedure.}
The participants were provided a textual description of several tasks to be completed.\footnote{We provide the full task description in the Appendix~\ref{app:usability}.}
Users were asked to complete a \texttt{Training} and \texttt{Prediction} action in a sentiment analysis scenario.
We provided both labeled test data and unlabeled training data again using the dataset by \cite{barbieri-etal-2020-tweeteval}.\footnote{Our focus is to evaluate the usability of the AdapterHub Playground application. Therefore, we did not require the participants to import the data on their own but rather provided them links to Google Docs containing the imported data.}
After completing the tasks, we asked the participants to complete a questionnaire targeting their experience with the tool.

\begin{figure}
    \centering
    \includegraphics[width=210pt]{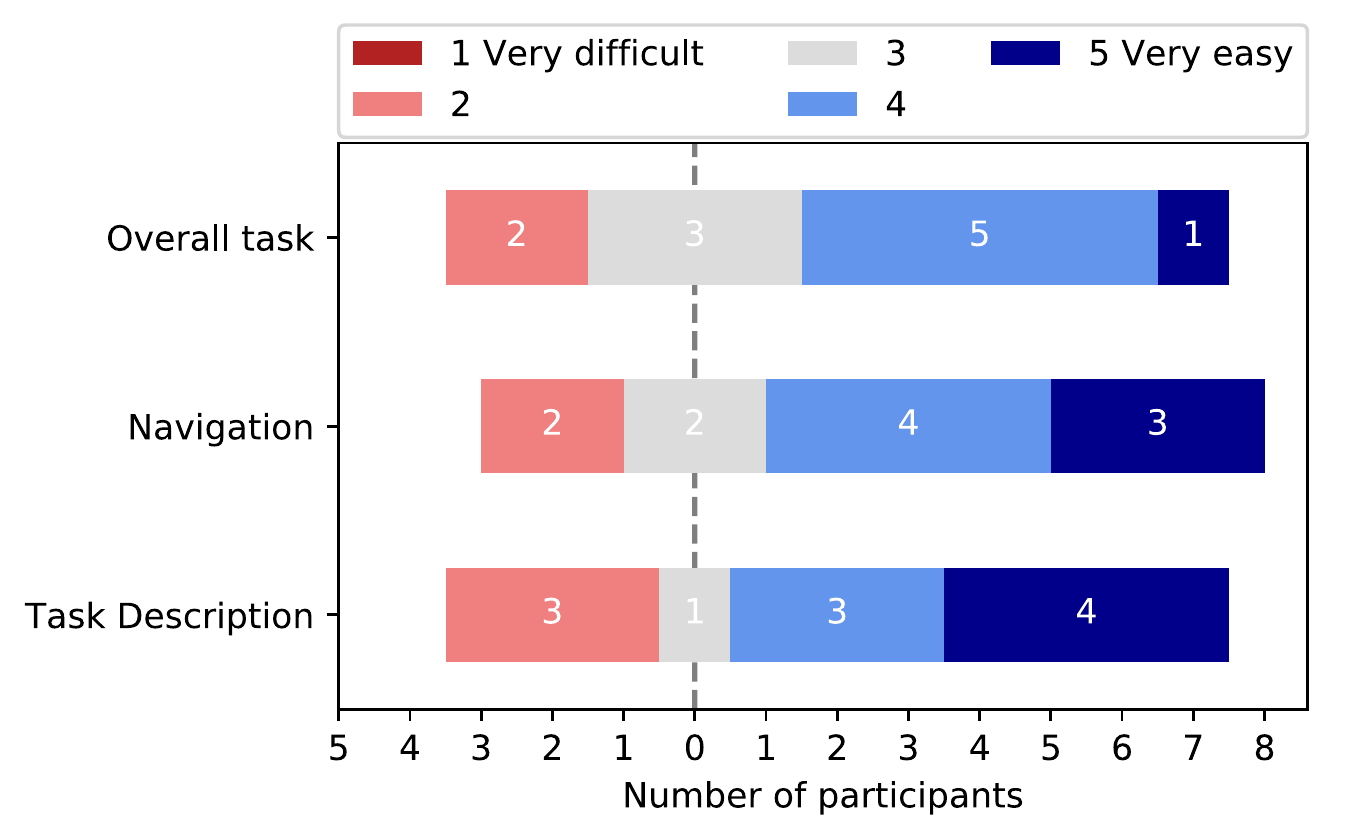}
    \caption{Participants' estimation of difficulty.}
    \label{fig:diff_experience}
    \vspace{-1em}
\end{figure}

\vspace{1mm}
\noindent\textbf{Results}
The participants were asked to assess the difficulties they faced on a five-point Likert scale, specifically, their experience with the overall task, the navigation of the application, and the difficulty of the task description (see Figure~\ref{fig:diff_experience}).
The majority of participants found the task and the navigation of the application to be simple.

Three participants found the task description difficult to understand.
We note here that the task description did not explain each individual navigation step in the application. This was designed on purpose - both to reduce the reading volume of the task description and to evaluate the accessibility of each feature of the application.

We further asked the participants about the difficulty of each individual task they had to solve, i.e. prediction, annotation, and training, on a five-point Likert scale ranging \textit{very difficult} (1) to \textit{very easy} (5).
Participants had the least trouble with the prediction action ($91\%$ voted either category 5 or 4; none voted category 1 or 2).
Despite the training action being technically similar to the prediction action, participants perceived it as more difficult with only $64\%$ selecting easier categories (4 and 5) and $27\%$ of the participants being undecided (category 3).
This is most likely due to some participants having issues finding the downloadable zip file which required opening the action detail page after training (we received this information as feedback in a free-answer form).

\section{Conclusion and Future Work}
\label{sec:conclusion}
Open-access dissemination of SotA NLP models has accelerated their use and popularity, yet
the required technical proficiency to apply them remains a limiting factor for their democratisation.
To mitigate this, we introduced the AdapterHub Playground application which provides an easy-to-use web interface for no-code access to light-weight adapters for text classification tasks.
Our system is built on-top of the open-source AdapterHub library and uses a dynamic code generation approach.
We demonstrated the features of the application using two exemplary use-case scenarios and evaluated its usability in a user study with researchers from communication sciences.
In addition to providing execution on third-party hardware, we also enabled a self-hosted computational instance.

As future work, we plan to extend the application with dynamic user control over all hyperparameter specifications in the expert mode.
To support users in efficient sampling of profitable training instances, we plan to investigate the integration of active learning methods~\cite{yuan-etal-2020-cold}.
A running instance of our tool can be found under \url{https://adapter-hub.github.io/playground}.
and the open-source code\footnote{Licensed under Apache License 2.0.} under \url{https://github.com/Adapter-Hub/playground}.

\section*{Broader Impact Statement}
\paragraph{Intended Use}

Our proposed application can be used in several ways and by different audiences.
First of all, it allows evaluating the performance of already existing fine-tuned adapters for various prominent text classification tasks on holdout data, possibly from another domain.
Further, one can provide annotated data for any of the supported tasks and continue training the corresponding adapter.
Training procedures can be repeated using different hyperparameters to investigate the effect of those on the prediction performance.
This makes our application interesting for both our target group, i.e. researchers outside of NLP using text classification methods, as well as NLP researchers interested in comparing various adapter models without setting up the required codebase to do so.

\paragraph{Possible Risks}

Primarily, the goal of our application is to lower the technical entry barrier for users interested in using state-of-the-art text classification models.
These users usually also lack the expertise to evaluate all aspects of the language understanding capabilities of such a model, as compared to researchers from within the NLP domain.
Rightfully, one can argue that publishing such an application increases the opportunities to develop more \textit{bad} black box models, caused by limited evaluation and missing expertise.
This can lead to severe misjudgements if conclusions are drawn based on predictions of such a model.

While we cannot eliminate this risk, we would like to raise some points which, in our opinion, put it into perspective with regard to the benefits of having such an application.

From a broader perspective, the AdapterHub Playground contributes to the democratization of access to the latest NLP research by simplifying the process of applying language model adapters for training and prediction.
This is especially helpful for interdisciplinary research where the applied text classification tools often rely on outdated methods~\citep{stoll2020detecting} or off-the-shelf tools~\citep{sen-etal-2020-reliability}. 
As a consequence, details about the model architecture, training procedure or out-of-domain performance are mostly omitted. 
While this does not imply low performance on hold-out data per se, it limits the possibilities for model evaluation and demands a certain level of trust from the end user.
In many cases, adapting the model to the target domain is not possible or requires some technical proficiency.
In addition, these models are often trained once-and-for-all while our framework allows for an interactive approach to evaluate model performance and offers the rich variety of pretrained adapters being available from the community-driven AdapterHub.

Further, we argue that advancements in NLP research should be made available to the researchers most profiting from them as soon as possible - not only for the sake of accelerating research outside of NLP but also to enable a feedback loop informing NLP researchers about the shortcomings of such models.
While the generalization capabilities of state-of-the-art language models are subject to increased scrutiny within NLP~\citep{sanchez-etal-2018-behavior, gururangan-etal-2020-dont, tu-etal-2020-empirical}, the datasets and tasks to test them often originate from within the same community, thereby introducing a selection bias~\citep{ramponi-plank-2020-neural}.
By enabling interdisciplinary researchers to evaluate NLP models without the technical barriers involved, we are able to gain more insights about the robustness and out-of-domain performance of these models.
Our application is a first step into this direction.

\section*{Acknowledgements}
This work has been supported by the German Research Foundation (DFG) as part of the Research Training Group KRITIS No. GRK 2222/2 and by the LOEWE initiative (Hesse, Germany) within the emergenCITY center. We want to thank the participants who volunteered to participate in our user study as well as Luke Bates and Ilia Kuznetsov for their valuable feedback.

\bibliography{anthology,acl2022_demo}

\begin{thebibliography}{53}
\expandafter\ifx\csname natexlab\endcsname\relax\def\natexlab#1{#1}\fi

\bibitem[{Agirre et~al.(2012)Agirre, Cer, Diab, and
  Gonzalez-Agirre}]{agirre-etal-2012-semeval}
Eneko Agirre, Daniel Cer, Mona Diab, and Aitor Gonzalez-Agirre. 2012.
\newblock \href {https://www.aclweb.org/anthology/S12-1051} {{S}em{E}val-2012
  task 6: A pilot on semantic textual similarity}.
\newblock In \emph{*{SEM} 2012: The First Joint Conference on Lexical and
  Computational Semantics {--} Volume 1: Proceedings of the main conference and
  the shared task, and Volume 2: Proceedings of the Sixth International
  Workshop on Semantic Evaluation ({S}em{E}val 2012)}, pages 385--393,
  Montr{\'e}al, Canada. Association for Computational Linguistics.

\bibitem[{Akbik et~al.(2019)Akbik, Bergmann, Blythe, Rasul, Schweter, and
  Vollgraf}]{akbik-etal-2019-flair}
Alan Akbik, Tanja Bergmann, Duncan Blythe, Kashif Rasul, Stefan Schweter, and
  Roland Vollgraf. 2019.
\newblock \href {https://doi.org/10.18653/v1/N19-4010} {{FLAIR}: An easy-to-use
  framework for state-of-the-art {NLP}}.
\newblock In \emph{Proceedings of the 2019 Conference of the North {A}merican
  Chapter of the Association for Computational Linguistics (Demonstrations)},
  pages 54--59, Minneapolis, Minnesota. Association for Computational
  Linguistics.

\bibitem[{Bapna and Firat(2019)}]{bapna-firat-2019-simple}
Ankur Bapna and Orhan Firat. 2019.
\newblock \href {https://doi.org/10.18653/v1/D19-1165} {Simple, scalable
  adaptation for neural machine translation}.
\newblock In \emph{Proceedings of the 2019 Conference on Empirical Methods in
  Natural Language Processing and the 9th International Joint Conference on
  Natural Language Processing (EMNLP-IJCNLP)}, pages 1538--1548, Hong Kong,
  China. Association for Computational Linguistics.

\bibitem[{Barbieri et~al.(2020)Barbieri, Camacho-Collados, Espinosa~Anke, and
  Neves}]{barbieri-etal-2020-tweeteval}
Francesco Barbieri, Jose Camacho-Collados, Luis Espinosa~Anke, and Leonardo
  Neves. 2020.
\newblock \href {https://doi.org/10.18653/v1/2020.findings-emnlp.148}
  {{T}weet{E}val: Unified benchmark and comparative evaluation for tweet
  classification}.
\newblock In \emph{Findings of the Association for Computational Linguistics:
  EMNLP 2020}, pages 1644--1650, Online. Association for Computational
  Linguistics.

\bibitem[{Beck et~al.(2021)Beck, Lee, Viehmann, Maurer, Quiring, and
  Gurevych}]{beck2021investigating}
Tilman Beck, Ji-Ung Lee, Christina Viehmann, Marcus Maurer, Oliver Quiring, and
  Iryna Gurevych. 2021.
\newblock \href {https://doi.org/10.18653/v1/2021.acl-long.1} {Investigating
  label suggestions for opinion mining in {G}erman covid-19 social media}.
\newblock In \emph{Proceedings of the 59th Annual Meeting of the Association
  for Computational Linguistics and the 11th International Joint Conference on
  Natural Language Processing (Volume 1: Long Papers)}, pages 1--13, Online.
  Association for Computational Linguistics.

\bibitem[{Boumans and Trilling(2016)}]{boumans2016taking}
Jelle~W Boumans and Damian Trilling. 2016.
\newblock \href {https://doi.org/10.1080/21670811.2015.1096598} {Taking stock
  of the toolkit: An overview of relevant automated content analysis approaches
  and techniques for digital journalism scholars}.
\newblock \emph{Digital journalism}, 4(1):8--23.

\bibitem[{Cer et~al.(2017)Cer, Diab, Agirre, Lopez-Gazpio, and
  Specia}]{cer-etal-2017-semeval}
Daniel Cer, Mona Diab, Eneko Agirre, I{\~n}igo Lopez-Gazpio, and Lucia Specia.
  2017.
\newblock \href {https://doi.org/10.18653/v1/S17-2001} {{S}em{E}val-2017 task
  1: Semantic textual similarity multilingual and crosslingual focused
  evaluation}.
\newblock In \emph{Proceedings of the 11th International Workshop on Semantic
  Evaluation ({S}em{E}val-2017)}, pages 1--14, Vancouver, Canada. Association
  for Computational Linguistics.

\bibitem[{Devlin et~al.(2019)Devlin, Chang, Lee, and
  Toutanova}]{devlin-etal-2019-bert}
Jacob Devlin, Ming-Wei Chang, Kenton Lee, and Kristina Toutanova. 2019.
\newblock \href {https://doi.org/10.18653/v1/N19-1423} {{BERT}: Pre-training of
  deep bidirectional transformers for language understanding}.
\newblock In \emph{Proceedings of the 2019 Conference of the North {A}merican
  Chapter of the Association for Computational Linguistics: Human Language
  Technologies, Volume 1 (Long and Short Papers)}, pages 4171--4186,
  Minneapolis, Minnesota. Association for Computational Linguistics.

\bibitem[{Dolan and Brockett(2005)}]{dolan-brockett-2005-automatically}
William~B. Dolan and Chris Brockett. 2005.
\newblock \href {https://www.aclweb.org/anthology/I05-5002} {Automatically
  constructing a corpus of sentential paraphrases}.
\newblock In \emph{Proceedings of the Third International Workshop on
  Paraphrasing ({IWP}2005)}.

\bibitem[{Efron and Tibshirani(1994)}]{efron1994introduction}
Bradley Efron and Robert~J Tibshirani. 1994.
\newblock \emph{An introduction to the bootstrap}.
\newblock CRC press.

\bibitem[{Grimmer and Stewart(2013)}]{grimmer2013text}
Justin Grimmer and Brandon~M Stewart. 2013.
\newblock \href {https://doi.org/10.1093/pan/mps028} {Text as data: {T}he
  promise and pitfalls of automatic content analysis methods for political
  texts}.
\newblock \emph{Political analysis}, 21(3):267--297.

\bibitem[{Gururangan et~al.(2020)Gururangan, Marasovi{\'c}, Swayamdipta, Lo,
  Beltagy, Downey, and Smith}]{gururangan-etal-2020-dont}
Suchin Gururangan, Ana Marasovi{\'c}, Swabha Swayamdipta, Kyle Lo, Iz~Beltagy,
  Doug Downey, and Noah~A. Smith. 2020.
\newblock \href {https://doi.org/10.18653/v1/2020.acl-main.740} {Don{'}t stop
  pretraining: Adapt language models to domains and tasks}.
\newblock In \emph{Proceedings of the 58th Annual Meeting of the Association
  for Computational Linguistics}, pages 8342--8360, Online. Association for
  Computational Linguistics.

\bibitem[{Han et~al.(2021)Han, Pang, and Wu}]{han-etal-2021-robust}
Wenjuan Han, Bo~Pang, and Ying~Nian Wu. 2021.
\newblock \href {https://doi.org/10.18653/v1/2021.acl-short.108} {Robust
  transfer learning with pretrained language models through adapters}.
\newblock In \emph{Proceedings of the 59th Annual Meeting of the Association
  for Computational Linguistics and the 11th International Joint Conference on
  Natural Language Processing (Volume 2: Short Papers)}, pages 854--861,
  Online. Association for Computational Linguistics.

\bibitem[{He et~al.(2021)He, Liu, Ye, Tan, Ding, Cheng, Low, Bing, and
  Si}]{he-etal-2021-effectiveness}
Ruidan He, Linlin Liu, Hai Ye, Qingyu Tan, Bosheng Ding, Liying Cheng, Jiawei
  Low, Lidong Bing, and Luo Si. 2021.
\newblock \href {https://doi.org/10.18653/v1/2021.acl-long.172} {On the
  effectiveness of adapter-based tuning for pretrained language model
  adaptation}.
\newblock In \emph{Proceedings of the 59th Annual Meeting of the Association
  for Computational Linguistics and the 11th International Joint Conference on
  Natural Language Processing (Volume 1: Long Papers)}, pages 2208--2222,
  Online. Association for Computational Linguistics.

\bibitem[{Houlsby et~al.(2019)Houlsby, Giurgiu, Jastrzebski, Morrone,
  De~Laroussilhe, Gesmundo, Attariyan, and Gelly}]{houlsby2019parameter}
Neil Houlsby, Andrei Giurgiu, Stanislaw Jastrzebski, Bruna Morrone, Quentin
  De~Laroussilhe, Andrea Gesmundo, Mona Attariyan, and Sylvain Gelly. 2019.
\newblock \href {http://proceedings.mlr.press/v97/houlsby19a/houlsby19a.pdf}
  {Parameter-efficient transfer learning for {NLP}}.
\newblock In \emph{International Conference on Machine Learning}, pages
  2790--2799. PMLR.

\bibitem[{Hultman et~al.(2018)Hultman, McEwan, Pakhomov, Lindemann, Skube, and
  Melton}]{hultman2018usability}
Gretchen Hultman, Reed McEwan, Serguei Pakhomov, Elizabeth Lindemann, Steven
  Skube, and Genevieve~B Melton. 2018.
\newblock \href {https://www.ncbi.nlm.nih.gov/pmc/articles/PMC5961783/}
  {Usability {E}valuation of an {U}nstructured {C}linical {D}ocument {Q}uery
  {T}ool for {R}esearchers}.
\newblock \emph{AMIA Summits on Translational Science Proceedings}, 2018:84.

\bibitem[{Lauscher et~al.(2020)Lauscher, Majewska, Ribeiro, Gurevych, Rozanov,
  and Glava{\v{s}}}]{lauscher-etal-2020-common}
Anne Lauscher, Olga Majewska, Leonardo F.~R. Ribeiro, Iryna Gurevych, Nikolai
  Rozanov, and Goran Glava{\v{s}}. 2020.
\newblock \href {https://doi.org/10.18653/v1/2020.deelio-1.5} {Common sense or
  world knowledge? investigating adapter-based knowledge injection into
  pretrained transformers}.
\newblock In \emph{Proceedings of Deep Learning Inside Out (DeeLIO): The First
  Workshop on Knowledge Extraction and Integration for Deep Learning
  Architectures}, pages 43--49, Online. Association for Computational
  Linguistics.

\bibitem[{Lei et~al.(2016)Lei, Joshi, Barzilay, Jaakkola, Tymoshenko,
  Moschitti, and M{\`a}rquez}]{lei-etal-2016-semi}
Tao Lei, Hrishikesh Joshi, Regina Barzilay, Tommi Jaakkola, Kateryna
  Tymoshenko, Alessandro Moschitti, and Llu{\'\i}s M{\`a}rquez. 2016.
\newblock \href {https://doi.org/10.18653/v1/N16-1153} {Semi-supervised
  question retrieval with gated convolutions}.
\newblock In \emph{Proceedings of the 2016 Conference of the North {A}merican
  Chapter of the Association for Computational Linguistics: Human Language
  Technologies}, pages 1279--1289, San Diego, California. Association for
  Computational Linguistics.

\bibitem[{Liu et~al.(2019)Liu, Ott, Goyal, Du, Joshi, Chen, Levy, Lewis,
  Zettlemoyer, and Stoyanov}]{liu2019roberta}
Yinhan Liu, Myle Ott, Naman Goyal, Jingfei Du, Mandar Joshi, Danqi Chen, Omer
  Levy, Mike Lewis, Luke Zettlemoyer, and Veselin Stoyanov. 2019.
\newblock \href {https://arxiv.org/pdf/1907.11692.pdf} {Roberta: A robustly
  optimized bert pretraining approach}.
\newblock \emph{arXiv preprint}.

\bibitem[{Maas et~al.(2011)Maas, Daly, Pham, Huang, Ng, and
  Potts}]{maas-etal-2011-learning}
Andrew~L. Maas, Raymond~E. Daly, Peter~T. Pham, Dan Huang, Andrew~Y. Ng, and
  Christopher Potts. 2011.
\newblock \href {https://www.aclweb.org/anthology/P11-1015} {Learning word
  vectors for sentiment analysis}.
\newblock In \emph{Proceedings of the 49th Annual Meeting of the Association
  for Computational Linguistics: Human Language Technologies}, pages 142--150,
  Portland, Oregon, USA. Association for Computational Linguistics.

\bibitem[{Mohammad et~al.(2016)Mohammad, Kiritchenko, Sobhani, Zhu, and
  Cherry}]{mohammad-etal-2016-dataset}
Saif Mohammad, Svetlana Kiritchenko, Parinaz Sobhani, Xiaodan Zhu, and Colin
  Cherry. 2016.
\newblock \href {https://www.aclweb.org/anthology/L16-1623} {A dataset for
  detecting stance in tweets}.
\newblock In \emph{Proceedings of the Tenth International Conference on
  Language Resources and Evaluation ({LREC}'16)}, pages 3945--3952,
  Portoro{\v{z}}, Slovenia. European Language Resources Association (ELRA).

\bibitem[{Pang and Lee(2005)}]{pang-lee-2005-seeing}
Bo~Pang and Lillian Lee. 2005.
\newblock \href {https://doi.org/10.3115/1219840.1219855} {Seeing stars:
  Exploiting class relationships for sentiment categorization with respect to
  rating scales}.
\newblock In \emph{Proceedings of the 43rd Annual Meeting of the Association
  for Computational Linguistics ({ACL}{'}05)}, pages 115--124, Ann Arbor,
  Michigan. Association for Computational Linguistics.

\bibitem[{Pedregosa et~al.(2011)Pedregosa, Varoquaux, Gramfort, Michel,
  Thirion, Grisel, Blondel, Prettenhofer, Weiss, Dubourg, Vanderplas, Passos,
  Cournapeau, Brucher, Perrot, and Duchesnay}]{scikit-learn}
F.~Pedregosa, G.~Varoquaux, A.~Gramfort, V.~Michel, B.~Thirion, O.~Grisel,
  M.~Blondel, P.~Prettenhofer, R.~Weiss, V.~Dubourg, J.~Vanderplas, A.~Passos,
  D.~Cournapeau, M.~Brucher, M.~Perrot, and E.~Duchesnay. 2011.
\newblock \href {https://dl.acm.org/doi/pdf/10.5555/1953048.2078195}
  {Scikit-learn: Machine learning in {P}ython}.
\newblock \emph{Journal of Machine Learning Research}, 12:2825--2830.

\bibitem[{Pfeiffer et~al.(2021{\natexlab{a}})Pfeiffer, Kamath, R{\"u}ckl{\'e},
  Cho, and Gurevych}]{pfeiffer-etal-2021-adapterfusion}
Jonas Pfeiffer, Aishwarya Kamath, Andreas R{\"u}ckl{\'e}, Kyunghyun Cho, and
  Iryna Gurevych. 2021{\natexlab{a}}.
\newblock \href {https://www.aclweb.org/anthology/2021.eacl-main.39}
  {{A}dapter{F}usion: Non-destructive task composition for transfer learning}.
\newblock In \emph{Proceedings of the 16th Conference of the European Chapter
  of the Association for Computational Linguistics: Main Volume}, pages
  487--503, Online. Association for Computational Linguistics.

\bibitem[{Pfeiffer et~al.(2020{\natexlab{a}})Pfeiffer, R{\"u}ckl{\'e}, Poth,
  Kamath, Vuli{\'c}, Ruder, Cho, and Gurevych}]{pfeiffer-etal-2020-adapterhub}
Jonas Pfeiffer, Andreas R{\"u}ckl{\'e}, Clifton Poth, Aishwarya Kamath, Ivan
  Vuli{\'c}, Sebastian Ruder, Kyunghyun Cho, and Iryna Gurevych.
  2020{\natexlab{a}}.
\newblock \href {https://doi.org/10.18653/v1/2020.emnlp-demos.7}
  {{A}dapter{H}ub: A framework for adapting transformers}.
\newblock In \emph{Proceedings of the 2020 Conference on Empirical Methods in
  Natural Language Processing: System Demonstrations}, pages 46--54, Online.
  Association for Computational Linguistics.

\bibitem[{Pfeiffer et~al.(2020{\natexlab{b}})Pfeiffer, Vuli{\'c}, Gurevych, and
  Ruder}]{pfeiffer-etal-2020-mad}
Jonas Pfeiffer, Ivan Vuli{\'c}, Iryna Gurevych, and Sebastian Ruder.
  2020{\natexlab{b}}.
\newblock \href {https://doi.org/10.18653/v1/2020.emnlp-main.617} {{MAD-X}:
  {A}n {A}dapter-{B}ased {F}ramework for {M}ulti-{T}ask {C}ross-{L}ingual
  {T}ransfer}.
\newblock In \emph{Proceedings of the 2020 Conference on Empirical Methods in
  Natural Language Processing (EMNLP)}, pages 7654--7673, Online. Association
  for Computational Linguistics.

\bibitem[{Pfeiffer et~al.(2021{\natexlab{b}})Pfeiffer, Vuli{\'c}, Gurevych, and
  Ruder}]{pfeiffer-etal-2021-unks}
Jonas Pfeiffer, Ivan Vuli{\'c}, Iryna Gurevych, and Sebastian Ruder.
  2021{\natexlab{b}}.
\newblock \href {https://doi.org/10.18653/v1/2021.emnlp-main.800} {{UNK}s
  everywhere: {A}dapting multilingual language models to new scripts}.
\newblock In \emph{Proceedings of the 2021 Conference on Empirical Methods in
  Natural Language Processing}, pages 10186--10203, Online and Punta Cana,
  Dominican Republic. Association for Computational Linguistics.

\bibitem[{Phang et~al.(2018)Phang, F{\'e}vry, and Bowman}]{phang2018sentence}
Jason Phang, Thibault F{\'e}vry, and Samuel~R Bowman. 2018.
\newblock \href {https://arxiv.org/pdf/1811.01088.pdf} {Sentence encoders on
  stilts: {S}upplementary training on intermediate labeled-data tasks}.
\newblock \emph{arXiv preprint}.

\bibitem[{Philip et~al.(2020)Philip, Berard, Gall{\'e}, and
  Besacier}]{philip-etal-2020-monolingual}
Jerin Philip, Alexandre Berard, Matthias Gall{\'e}, and Laurent Besacier. 2020.
\newblock \href {https://doi.org/10.18653/v1/2020.emnlp-main.361} {Monolingual
  adapters for zero-shot neural machine translation}.
\newblock In \emph{Proceedings of the 2020 Conference on Empirical Methods in
  Natural Language Processing (EMNLP)}, pages 4465--4470, Online. Association
  for Computational Linguistics.

\bibitem[{Poth et~al.(2021)Poth, Pfeiffer, R{\"u}ckl{\'e}, and
  Gurevych}]{poth-etal-2021-pre}
Clifton Poth, Jonas Pfeiffer, Andreas R{\"u}ckl{\'e}, and Iryna Gurevych. 2021.
\newblock \href {https://doi.org/10.18653/v1/2021.emnlp-main.827} {{W}hat to
  pre-train on? {E}fficient intermediate task selection}.
\newblock In \emph{Proceedings of the 2021 Conference on Empirical Methods in
  Natural Language Processing}, pages 10585--10605, Online and Punta Cana,
  Dominican Republic. Association for Computational Linguistics.

\bibitem[{Pruksachatkun et~al.(2020)Pruksachatkun, Phang, Liu, Htut, Zhang,
  Pang, Vania, Kann, and Bowman}]{pruksachatkun-etal-2020-intermediate}
Yada Pruksachatkun, Jason Phang, Haokun Liu, Phu~Mon Htut, Xiaoyi Zhang,
  Richard~Yuanzhe Pang, Clara Vania, Katharina Kann, and Samuel~R. Bowman.
  2020.
\newblock \href {https://doi.org/10.18653/v1/2020.acl-main.467}
  {Intermediate-task transfer learning with pretrained language models: When
  and why does it work?}
\newblock In \emph{Proceedings of the 58th Annual Meeting of the Association
  for Computational Linguistics}, pages 5231--5247, Online. Association for
  Computational Linguistics.

\bibitem[{Ramponi and Plank(2020)}]{ramponi-plank-2020-neural}
Alan Ramponi and Barbara Plank. 2020.
\newblock \href {https://doi.org/10.18653/v1/2020.coling-main.603} {Neural
  unsupervised domain adaptation in {NLP}{---}{A} survey}.
\newblock In \emph{Proceedings of the 28th International Conference on
  Computational Linguistics}, pages 6838--6855, Barcelona, Spain (Online).
  International Committee on Computational Linguistics.

\bibitem[{Rebuffi et~al.(2017)Rebuffi, Bilen, and Vedaldi}]{NIPS2017_e7b24b11}
Sylvestre-Alvise Rebuffi, Hakan Bilen, and Andrea Vedaldi. 2017.
\newblock \href
  {https://proceedings.neurips.cc/paper/2017/file/e7b24b112a44fdd9ee93bdf998c6ca0e-Paper.pdf}
  {Learning multiple visual domains with residual adapters}.
\newblock In \emph{Advances in Neural Information Processing Systems},
  volume~30. Curran Associates, Inc.

\bibitem[{Reimers and Gurevych(2019)}]{reimers-gurevych-2019-sentence}
Nils Reimers and Iryna Gurevych. 2019.
\newblock \href {https://doi.org/10.18653/v1/D19-1410} {Sentence-{BERT}:
  Sentence embeddings using {S}iamese {BERT}-networks}.
\newblock In \emph{Proceedings of the 2019 Conference on Empirical Methods in
  Natural Language Processing and the 9th International Joint Conference on
  Natural Language Processing (EMNLP-IJCNLP)}, pages 3982--3992, Hong Kong,
  China. Association for Computational Linguistics.

\bibitem[{Ribeiro et~al.(2021)Ribeiro, Zhang, and
  Gurevych}]{ribeiro-etal-2021-structural}
Leonardo F.~R. Ribeiro, Yue Zhang, and Iryna Gurevych. 2021.
\newblock \href {https://doi.org/10.18653/v1/2021.emnlp-main.351} {Structural
  adapters in pretrained language models for {AMR}-to-{T}ext generation}.
\newblock In \emph{Proceedings of the 2021 Conference on Empirical Methods in
  Natural Language Processing}, pages 4269--4282, Online and Punta Cana,
  Dominican Republic. Association for Computational Linguistics.

\bibitem[{Rosenthal et~al.(2017)Rosenthal, Farra, and
  Nakov}]{rosenthal-etal-2017-semeval}
Sara Rosenthal, Noura Farra, and Preslav Nakov. 2017.
\newblock \href {https://doi.org/10.18653/v1/S17-2088} {{S}em{E}val-2017 task
  4: Sentiment analysis in {T}witter}.
\newblock In \emph{Proceedings of the 11th International Workshop on Semantic
  Evaluation ({S}em{E}val-2017)}, pages 502--518, Vancouver, Canada.
  Association for Computational Linguistics.

\bibitem[{R{\"u}ckl{\'e} et~al.(2021{\natexlab{a}})R{\"u}ckl{\'e}, Geigle,
  Glockner, Beck, Pfeiffer, Reimers, and
  Gurevych}]{ruckle-etal-2021-adapterdrop}
Andreas R{\"u}ckl{\'e}, Gregor Geigle, Max Glockner, Tilman Beck, Jonas
  Pfeiffer, Nils Reimers, and Iryna Gurevych. 2021{\natexlab{a}}.
\newblock \href {https://doi.org/10.18653/v1/2021.emnlp-main.626}
  {{AdapterDrop}: {O}n the efficiency of adapters in transformers}.
\newblock In \emph{Proceedings of the 2021 Conference on Empirical Methods in
  Natural Language Processing}, pages 7930--7946, Online and Punta Cana,
  Dominican Republic. Association for Computational Linguistics.

\bibitem[{R{\"u}ckl{\'e} et~al.(2021{\natexlab{b}})R{\"u}ckl{\'e}, Geigle,
  Glockner, Beck, Pfeiffer, Reimers, and Gurevych}]{ruckle2020adapterdrop}
Andreas R{\"u}ckl{\'e}, Gregor Geigle, Max Glockner, Tilman Beck, Jonas
  Pfeiffer, Nils Reimers, and Iryna Gurevych. 2021{\natexlab{b}}.
\newblock \href {https://doi.org/10.18653/v1/2021.emnlp-main.626}
  {{AdapterDrop}: {O}n the efficiency of adapters in transformers}.
\newblock In \emph{Proceedings of the 2021 Conference on Empirical Methods in
  Natural Language Processing}, pages 7930--7946, Online and Punta Cana,
  Dominican Republic. Association for Computational Linguistics.

\bibitem[{R{\"u}ckl{\'e} et~al.(2020)R{\"u}ckl{\'e}, Pfeiffer, and
  Gurevych}]{ruckle-etal-2020-multicqa}
Andreas R{\"u}ckl{\'e}, Jonas Pfeiffer, and Iryna Gurevych. 2020.
\newblock \href {https://doi.org/10.18653/v1/2020.emnlp-main.194}
  {{M}ulti{CQA}: Zero-shot transfer of self-supervised text matching models on
  a massive scale}.
\newblock In \emph{Proceedings of the 2020 Conference on Empirical Methods in
  Natural Language Processing (EMNLP)}, pages 2471--2486, Online. Association
  for Computational Linguistics.

\bibitem[{Sanchez et~al.(2018)Sanchez, Mitchell, and
  Riedel}]{sanchez-etal-2018-behavior}
Ivan Sanchez, Jeff Mitchell, and Sebastian Riedel. 2018.
\newblock \href {https://doi.org/10.18653/v1/N18-1179} {Behavior analysis of
  {NLI} models: Uncovering the influence of three factors on robustness}.
\newblock In \emph{Proceedings of the 2018 Conference of the North {A}merican
  Chapter of the Association for Computational Linguistics: Human Language
  Technologies, Volume 1 (Long Papers)}, pages 1975--1985, New Orleans,
  Louisiana. Association for Computational Linguistics.

\bibitem[{Schiller et~al.(2021)Schiller, Daxenberger, and
  Gurevych}]{schiller2021stance}
Benjamin Schiller, Johannes Daxenberger, and Iryna Gurevych. 2021.
\newblock \href {https://doi.org/10.1007/s13218-021-00714-w} {Stance detection
  benchmark: How robust is your stance detection?}
\newblock \emph{KI-K{\"u}nstliche Intelligenz}, pages 1--13.

\bibitem[{Sen et~al.(2020)Sen, Fl{\"o}ck, and
  Wagner}]{sen-etal-2020-reliability}
Indira Sen, Fabian Fl{\"o}ck, and Claudia Wagner. 2020.
\newblock \href {https://doi.org/10.18653/v1/2020.emnlp-main.110} {On the
  reliability and validity of detecting approval of political actors in
  tweets}.
\newblock In \emph{Proceedings of the 2020 Conference on Empirical Methods in
  Natural Language Processing (EMNLP)}, pages 1413--1426, Online. Association
  for Computational Linguistics.

\bibitem[{Socher et~al.(2013)Socher, Perelygin, Wu, Chuang, Manning, Ng, and
  Potts}]{socher-etal-2013-recursive}
Richard Socher, Alex Perelygin, Jean Wu, Jason Chuang, Christopher~D. Manning,
  Andrew Ng, and Christopher Potts. 2013.
\newblock \href {https://www.aclweb.org/anthology/D13-1170} {Recursive deep
  models for semantic compositionality over a sentiment treebank}.
\newblock In \emph{Proceedings of the 2013 Conference on Empirical Methods in
  Natural Language Processing}, pages 1631--1642, Seattle, Washington, USA.
  Association for Computational Linguistics.

\bibitem[{Stickland and Murray(2019)}]{pmlr-v97-stickland19a}
Asa~Cooper Stickland and Iain Murray. 2019.
\newblock \href {http://proceedings.mlr.press/v97/stickland19a.html} {{BERT}
  and {PAL}s: {P}rojected {A}ttention {L}ayers for {E}fficient {A}daptation in
  {M}ulti-{T}ask {L}earning}.
\newblock In \emph{Proceedings of the 36th International Conference on Machine
  Learning}, volume~97 of \emph{Proceedings of Machine Learning Research},
  pages 5986--5995. PMLR.

\bibitem[{Stoll et~al.(2020)Stoll, Ziegele, and Quiring}]{stoll2020detecting}
Anke Stoll, Marc Ziegele, and Oliver Quiring. 2020.
\newblock \href {https://computationalcommunication.org/ccr/article/view/19}
  {Detecting impoliteness and incivility in online discussions: Classification
  approaches for german user comments}.
\newblock \emph{Computational Communication Research}, 2(1):109--134.

\bibitem[{Tu et~al.(2020)Tu, Lalwani, Gella, and He}]{tu-etal-2020-empirical}
Lifu Tu, Garima Lalwani, Spandana Gella, and He~He. 2020.
\newblock \href {https://doi.org/10.1162/tacl_a_00335} {An empirical study on
  robustness to spurious correlations using pre-trained language models}.
\newblock \emph{Transactions of the Association for Computational Linguistics},
  8:621--633.

\bibitem[{{\"U}st{\"u}n et~al.(2020){\"U}st{\"u}n, Bisazza, Bouma, and van
  Noord}]{ustun-etal-2020-udapter}
Ahmet {\"U}st{\"u}n, Arianna Bisazza, Gosse Bouma, and Gertjan van Noord. 2020.
\newblock \href {https://doi.org/10.18653/v1/2020.emnlp-main.180} {{UD}apter:
  Language adaptation for truly {U}niversal {D}ependency parsing}.
\newblock In \emph{Proceedings of the 2020 Conference on Empirical Methods in
  Natural Language Processing (EMNLP)}, pages 2302--2315, Online. Association
  for Computational Linguistics.

\bibitem[{van Atteveldt and Peng(2021)}]{van2021computational}
Wouter van Atteveldt and Tai-Quan Peng. 2021.
\newblock \href
  {https://www.routledge.com/Computational-Methods-for-Communication-Science/Atteveldt-Peng/p/book/9780367536169}
  {\emph{Computational {M}ethods for {C}ommunication {S}cience}}.
\newblock Routledge.

\bibitem[{van Atteveldt et~al.(2021)van Atteveldt, van~der Velden, and
  Boukes}]{van2021validity}
Wouter van Atteveldt, Mariken~ACG van~der Velden, and Mark Boukes. 2021.
\newblock \href {https://doi.org/10.1080/19312458.2020.1869198} {The {V}alidity
  of {S}entiment {A}nalysis: {C}omparing {M}anual {A}nnotation,
  {C}rowd-{C}oding, {D}ictionary {A}pproaches, and {M}achine {L}earning
  {A}lgorithms}.
\newblock \emph{Communication Methods and Measures}, pages 1--20.

\bibitem[{Vidoni et~al.(2020)Vidoni, Vuli{\'c}, and
  Glava{\v{s}}}]{vidoni2020orthogonal}
Marko Vidoni, Ivan Vuli{\'c}, and Goran Glava{\v{s}}. 2020.
\newblock Orthogonal language and task adapters in zero-shot cross-lingual
  transfer.
\newblock \emph{arXiv preprint arXiv:2012.06460}.

\bibitem[{Wang et~al.(2021)Wang, Tang, Duan, Wei, Huang, Ji, Cao, Jiang, and
  Zhou}]{wang-etal-2021-k}
Ruize Wang, Duyu Tang, Nan Duan, Zhongyu Wei, Xuanjing Huang, Jianshu Ji,
  Guihong Cao, Daxin Jiang, and Ming Zhou. 2021.
\newblock \href {https://doi.org/10.18653/v1/2021.findings-acl.121}
  {{K-Adapter}: {I}nfusing {K}nowledge into {P}re-{T}rained {M}odels with
  {A}dapters}.
\newblock In \emph{Findings of the Association for Computational Linguistics:
  ACL-IJCNLP 2021}, pages 1405--1418, Online. Association for Computational
  Linguistics.

\bibitem[{Wolf et~al.(2020)Wolf, Debut, Sanh, Chaumond, Delangue, Moi, Cistac,
  Rault, Louf, Funtowicz, Davison, Shleifer, von Platen, Ma, Jernite, Plu, Xu,
  Le~Scao, Gugger, Drame, Lhoest, and Rush}]{wolf-etal-2020-transformers}
Thomas Wolf, Lysandre Debut, Victor Sanh, Julien Chaumond, Clement Delangue,
  Anthony Moi, Pierric Cistac, Tim Rault, Remi Louf, Morgan Funtowicz, Joe
  Davison, Sam Shleifer, Patrick von Platen, Clara Ma, Yacine Jernite, Julien
  Plu, Canwen Xu, Teven Le~Scao, Sylvain Gugger, Mariama Drame, Quentin Lhoest,
  and Alexander Rush. 2020.
\newblock \href {https://doi.org/10.18653/v1/2020.emnlp-demos.6} {Transformers:
  State-of-the-art natural language processing}.
\newblock In \emph{Proceedings of the 2020 Conference on Empirical Methods in
  Natural Language Processing: System Demonstrations}, pages 38--45, Online.
  Association for Computational Linguistics.

\bibitem[{Yuan et~al.(2020)Yuan, Lin, and Boyd-Graber}]{yuan-etal-2020-cold}
Michelle Yuan, Hsuan-Tien Lin, and Jordan Boyd-Graber. 2020.
\newblock \href {https://doi.org/10.18653/v1/2020.emnlp-main.637} {Cold-start
  active learning through self-supervised language modeling}.
\newblock In \emph{Proceedings of the 2020 Conference on Empirical Methods in
  Natural Language Processing (EMNLP)}, pages 7935--7948, Online. Association
  for Computational Linguistics.

\end{thebibliography}
\bibliographystyle{acl_natbib}

\clearpage
\appendix

\section{Frequently Asked Questions}
\label{app:qa}

\paragraph{What are the requirements to use the AdapterHub Playground?}

The most basic usage requirement is an up-to-date modern web browser\footnote{We tested the application using Desktop Firefox and Desktop Chrome.}.
To use the application without setting up your own computing instance, one needs to create a Kaggle account and download the API Token for login.
We provide information on setting up a local compute instance at \url{https://github.com/Adapter-Hub/playground}.
As we use GoogleSheet as data hosting platform, the user needs an active Google account.

If used for prediction, textual target data must be uploaded to a GoogleSheet and linked with a project within the application.
For training, each text must be additionally labelled according to the target task's label matching schema.
While we also provide information about each supported task on a separate page in the application, we expect the user to have a basic understanding of the procedure of the targeted task (e.g. \textit{Sentiment Analysis} is about predicting the sentiment tone of a given text).

\paragraph{How is a user able to identify label mismatch?}

For each supported task, we provide the necessary label matching information within the dialogue to create a new Action (e.g. in Figure \ref{fig:taskcreation}).
If the user-provided labels in the Google data sheet do not match the selected task or adapter architecture, an error message will be provided giving information about the indices of the mismatched data points.

\paragraph{How could a new user determine the task for their data?}

In general, this application is intended for users who know what type of predictions (i.e. the task) they want to apply on their data.
We provide support for a subset of (pairwise) text classification tasks from AdapterHub.ml with the goal to cover the most prominent ones used in interdisciplinary research.
However, we also provide basic information about each supported task on a separate page in the application.

\paragraph{What if my task is not supported in the AdapterHub Playground?}

In general, we can only provide support for the classification tasks which are covered on Adapterhub.
We have selected a subset of tasks which we deem to be of interest in interdisciplinary research (e.g. computational social science).
Integration of new tasks is possible by extending the application which requires some technical background in coding and web development.
If you are a researcher and lack the technical proficiency to do so, we encourage you to get into contact with us to find out if and how we can integrate your task.

\paragraph{Which pretrained adapter should be used?}

This is still an open research question and we refer to the literature for more details \citep{phang2018sentence, pruksachatkun-etal-2020-intermediate, poth-etal-2021-pre}.
However, there are some heuristics which can be followed.
Regarding adapters, it has been shown that domain similarity (e.g. training and test data are both from Twitter) and training dataset size (the more the better) can be indicators for good transfer performance~\citep{poth-etal-2021-pre}.

\paragraph{How should hyperparameters be set?}

Hyperparameter optimization for machine learning is a research field of its own and there is no one-size-fits-all solution to this.
Especially for users without experience in tuning ML models identifying reasonable hyperparameter values might seem rather arbitrary.

Currently, we support tuning the learning rate and the number of epochs.
In general, if the learning rate is high the training may not converge or even diverge. 
The changes in the weights might become too big such that the optimizer will not find optimal values.
A low learning rate is good, but the model will take more iterations to converge because steps towards the minimum of the loss function are tiny.
In practice it is good strategy to test different (high and low) learning rates to identify their effect on the model performance.

One epoch describes a full cycle through the entire training dataset.
A single epoch can sometimes be enough to improve performance significantly and training text classification adapters longer than for 10 epochs rarely provides substantial improvements. We recommend testing different numbers of epochs (between 2 and 5) to evaluate if longer training is beneficial for the task at hand.

\section{Training Details}
\label{app:training}

\begin{table*}[!htbp]
    \begin{tabular}{llll}
    \toprule
    Adapter & Task & Pretrained Language Model Identifier & Architecture \\ 
    \midrule
    \texttt{IMDB} & Sentiment Analysis & \texttt{distilbert-base-uncased} & Pfeiffer  \\
    \texttt{RT} & Sentiment Analysis & \texttt{distilbert-base-uncased} & Pfeiffer \\
    \texttt{SST-2} & Sentiment Analysis & \texttt{bert-base-uncased} & Houlsby \\
    \midrule
    \texttt{MRPC} & Semantic Textual Similarity & \texttt{bert-base-uncased} & Houlsby \\
    \texttt{QQP} & Semantic Textual Similarity & \texttt{bert-base-uncased} & Houlsby \\
    \bottomrule
    \end{tabular}
    \caption{Adapter architecture details for each specific task.}
    \label{tab:architectures}
\end{table*}

We did not perform any hyperparameter optimization for our experiments and used the default settings in the AdapterHub Playground application.
We adopted a learning rate of 1e-4 from related work~\cite{pfeiffer-etal-2020-adapterhub} and trained each adapter for three epochs.
In Table~\ref{tab:architectures} we provide the respective adapter architectures which were used for each specific adapter.

\section{Extensibility}
\label{app:extensibility}

Extending the AdapterHub Playground with a new text classification task requires adaptations to both the frontend and backend.

The repository supports a deployment workflow which will update a configuration file with all relevant information from the AdapterHub.
This enables that all tasks and their corresponding pretrained adapters (with a classification head) are potentially available within the AdapterHub Playground. 
The tasks for these adapters are filtered based on a predefined set of tasks which should be available to users of the application. 
Within the application, the filter list needs to be adapted such that the new task is not filtered during startup of the application. 
Additionally, the task name and its description need to be added to the frontend code as well as the label mapping information.
In the backend we need to add the label mapping and the list of supported tasks such that the evaluation computation is correct.

We provide the technical details within the code repository at \url{https://github.com/Adapter-Hub/playground}.

\section{Usability Study}
\label{app:usability}

\subsection{Participants}

\begin{figure}
    \centering
    \includegraphics[width=210pt]{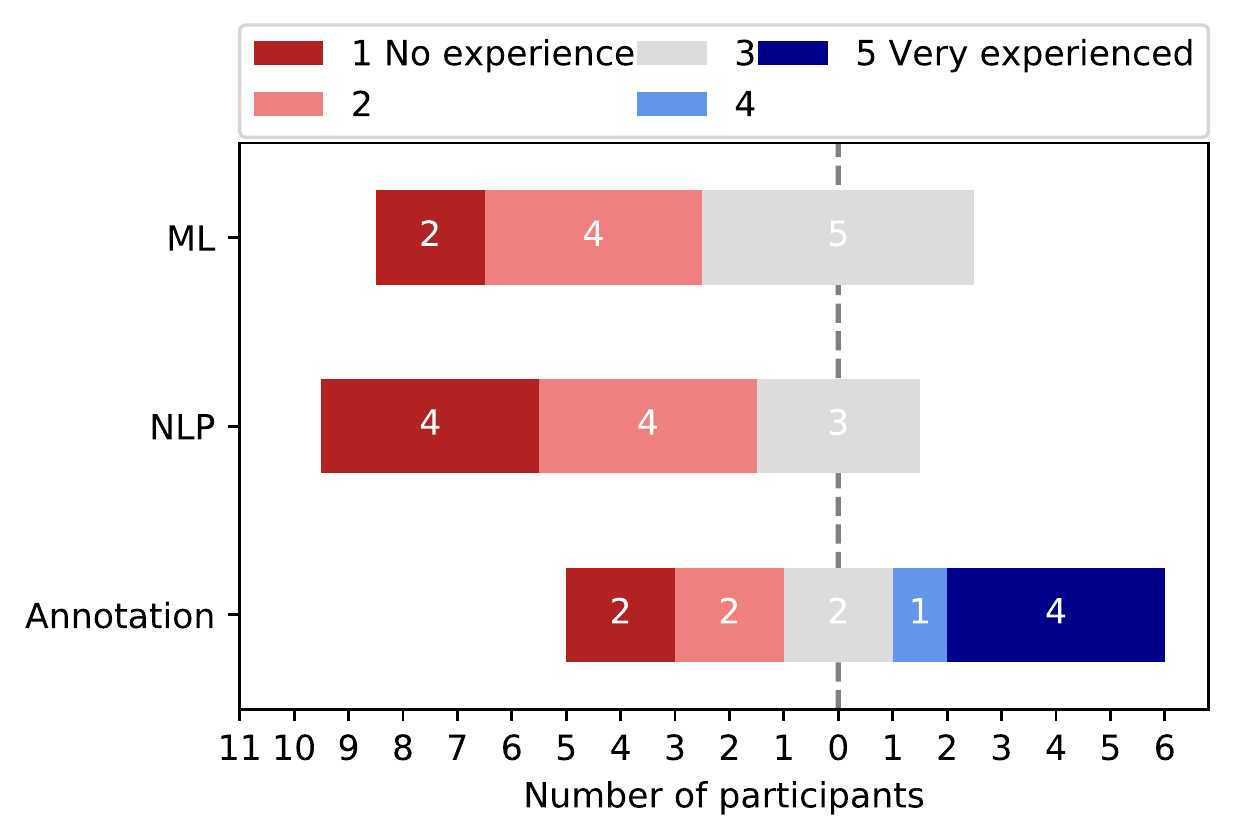}
    \caption{Participants' experience with underlying technical concepts.}
    \label{fig:study_experience}
\end{figure}

As can be seen in Figure~\ref{fig:study_experience}, most participants have only a basic understanding of the technical concepts related to machine learning or natural language processing.
However, it is likely they have experience with annotating data.
We further asked them if they can envision using the AdapterHub Playground application in their research.
Slightly more than half gave a positive answer ($54\%$) and the rest were undecided; no participant claimed they would never use our application.

Thus, we conclude that our participants belong to our target group.

\subsection{Instructions}

\begin{figure}[ht]
    \noindent
    \centering
    \includegraphics[page=2,width=0.75\paperwidth]{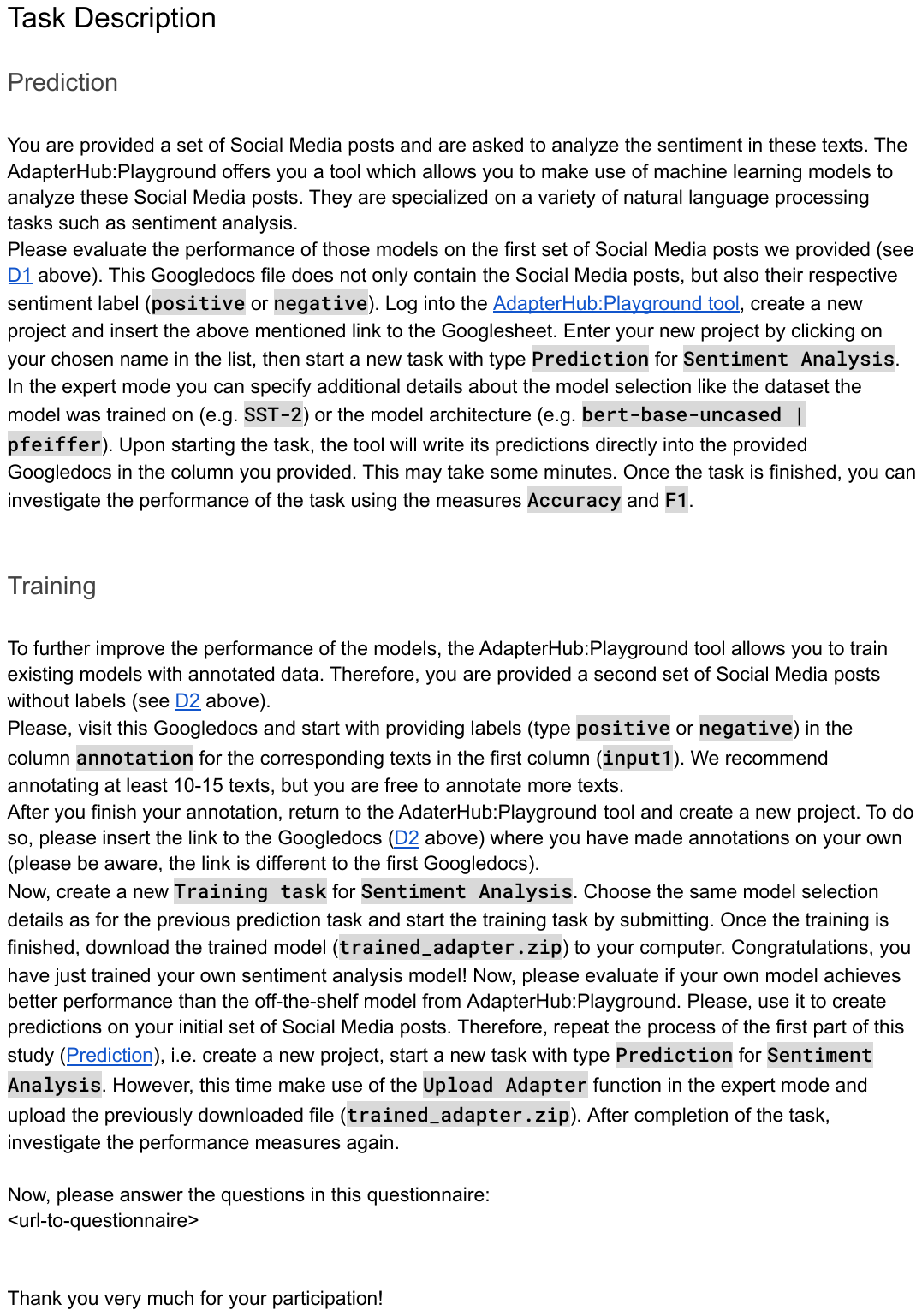}
    \caption{Instructions for the participants of the user study.}
    \label{fig:instructions}
\end{figure}

We provide the instructions for the usability study in Figure~\ref{fig:instructions}.

\clearpage

\begin{figure}[ht]
    \noindent
    \centering
    \includegraphics[page=1,width=0.75\paperwidth]{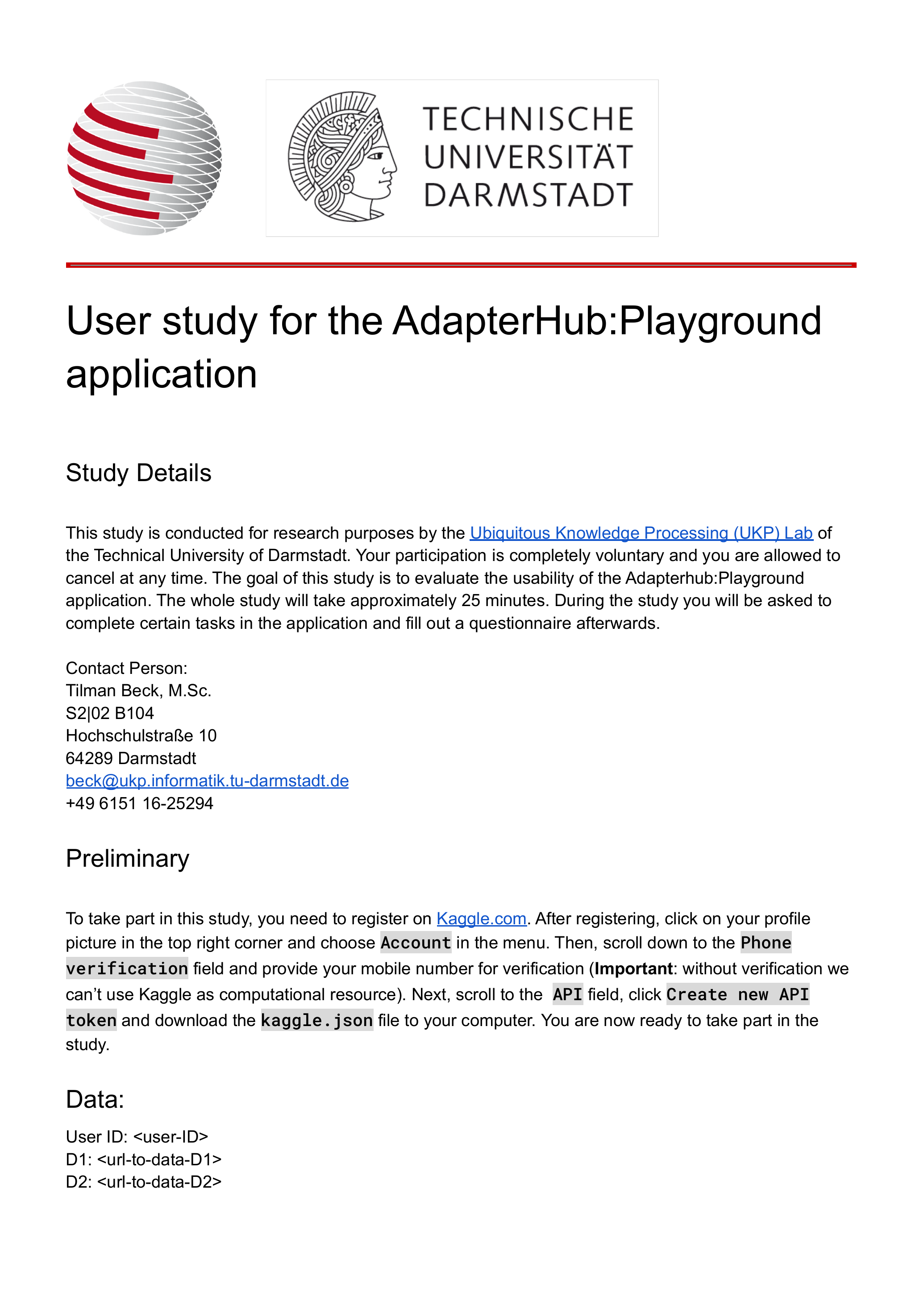}
\end{figure}

\clearpage

\clearpage

\end{document}